\documentclass{article}

 \usepackage[preprint]{neurips_2026}

\usepackage[utf8]{inputenc} 
\usepackage[T1]{fontenc}    
\usepackage{hyperref}       
\usepackage{url}            
\usepackage{booktabs}       
\usepackage{amsfonts}       
\usepackage{nicefrac}       
\usepackage{microtype}      
\usepackage{xcolor}         
\usepackage{hyperref}       
\usepackage{url}            
\usepackage{booktabs}       
\usepackage{amsfonts}       
\usepackage{nicefrac}       
\usepackage{microtype}      
\usepackage{lipsum}
\usepackage{fancyhdr}       
\usepackage{graphicx}       
\usepackage{natbib}

\usepackage{xcolor}
\usepackage{tcolorbox}
\tcbuselibrary{breakable}
\tcbuselibrary{breakable}
\usepackage{mathtools}
\usepackage{hyperref} 
\usepackage{cleveref}
\usepackage{algorithm}
\usepackage[noend]{algorithmic}
\usepackage{graphicx}
\usepackage{textcomp}
\usepackage{multirow}
\usepackage{xspace}
\usepackage{listings}
\usepackage{caption}
\usepackage{amsthm}
\usepackage{float}
\usepackage{soul}
\usepackage{upquote}
\usepackage{listings}
\usepackage{enumitem}
\usepackage{extarrows}
\usepackage{stmaryrd}
\usepackage{bbm}
\usepackage{subcaption}
\usepackage{makecell}
\usepackage{array}
\usepackage{dsfont}
\usepackage{colortbl}
\usepackage{tikz}
\usepackage{wrapfig}
\usepackage{amssymb}
\usepackage{diagbox}
\usepackage{bm}
\usepackage{fvextra}
\usepackage{thmtools}

\theoremstyle{plain}

\theoremstyle{definition}

\crefname{assumption}{assumption}{assumptions}
\Crefname{assumption}{Assumption}{Assumptions}

\newcommand{\tsf}[1]{\texttt{#1}}

\newcommand{\sys}{\tsf{DynMuon}\xspace}

\title{\sys: A Dynamic Spectral Shaping View of Muon
}

\author{Fangzhou Wu$^{1}$, Rikhav Shah$^{2,\dagger}$, Sandeep Silwal$^{1,\dagger}$, Qiuyi (Richard) Zhang$^{3,\dagger}$ \\
$^{1}$University of Wisconsin--Madison, $^{2}$MIT, $^{3}$Elorian AI\\
\texttt{fwu89@wisc.edu, rdshah@mit.edu, silwal@cs.wisc.edu, richard@elorian.ai} \\
}

\begin{document}

\maketitle

\begingroup
\renewcommand{\thefootnote}{\fnsymbol{footnote}}
\footnotetext[2]{The remaining authors are listed alphabetically.}
\endgroup

\begin{abstract}
In recent years, Muon has emerged as the dominant method for training large language models, and transformers more broadly. The essential difference, when compared to standard gradient descent methods, is to replace the usual update matrix $M=U\Sigma V^\top$ with its polar factor $UV^\top$.
In this work, we consider a class of Muon-like updates, where we replace the update $M$ with $U\Sigma^p V^\top$ for some parameter $p$. 
We call this a ``spectral-shaping'' operation, and develop a theory of how to pick $p$ which depends on (a) local curvature of the loss function, (b) noise stemming from stochastic gradients and label noise, and (c) training stage.
Our theory and experimentation reveal a previously overlooked behavior: positive $p$ helps early by emphasizing high-curvature directions and accelerating signal contraction, while mildly negative $p$ helps later by reallocating update strength toward low-curvature directions that still contain useful training signals.
Building on the insight, we propose \sys, an efficient dynamic spectral shaping method that schedules $p$ from positive to mildly negative over training.
Extensive experiments across model sizes, architectures, and training settings show that \sys consistently achieves lower validation loss than Muon, while requiring \textbf{10.6--26.5\%} fewer steps to reach the same target loss. 
Our code is available at \url{https://github.com/fzwark/DynMuon}.
\end{abstract}

\section{Introduction}
Muon has recently shown strong empirical performance for LLM training~\citep{jordan2024muon}. 
At a high level, for each matrix-valued model parameter, Muon forms a momentum-averaged gradient matrix $M=U\Sigma V^\top$ and replaces it with its polar factor $UV^\top$.
The resulting update preserves the singular directions while ``flattening'' the singular values, and has been shown to improve convergence and training stability across model scales~\citep{liu2025muonscalablellmtraining,ahn2025diondistributedorthonormalizedupdates}.
This flattening of singular values naturally suggests studying more general spectral transformations, which we call \emph{spectral shaping}, for matrix-valued optimization, and exploring how the relative weighting of spectral directions affects training dynamics.

Recent work has explored spectral shaping in a limited capacity by studying fixed power-law singular-value shapings~\citep{qi2026delvingmuonbeyonddeep,chen2026muon}, or transformations that suppress dominant spectral subspaces~\citep{huang2026spectrarethinkingoptimizersllms,yang2026prismstructuredoptimizationanisotropic}.
However, they still treat spectral shaping as static, seeking better fixed spectral transformations. 
Thus, it is not clear whether a single shaping rule is desirable as training dynamics evolve~\citep{qi2026delvingmuonbeyonddeep}. 
Relatedly, prior work also lacks a training-dynamics model of spectral shaping, and therefore does not characterize how the relative influence of different spectral directions evolves across training~\citep{huang2026spectrarethinkingoptimizersllms,yang2026prismstructuredoptimizationanisotropic}. 
This leaves a central question unresolved, which is the main focus of our paper: 
\vskip -0.3in
\begin{quote}
    \emph{How should Muon-style spectral shaping adapt across training stages, if at all?}
\end{quote}
\vskip -0.2in

\paragraph{Our Results}
To answer this question, we generalize Muon as one point in a power-law family of what we call \emph{spectral-shaping} operations, where $p$ is the  spectral exponent:
\vskip -0.2in
\begin{equation}
    D^{(p)} := U \Sigma^p V^\top, \qquad \Sigma^p = \mathrm{diag}(\sigma_1^p,\dots,\sigma_r^p),
\end{equation}
for a matrix-valued update $M=U\Sigma V^\top.$ This is demonstratively a very expressive family of operations: $p=-1$ gives an inverse-spectrum update, $p=0$ recovers Muon, and $p=1$ corresponds to the standard SGD-style updates. 
To understand training dynamics as a function of $p$, we develop a simple \emph{noise-aware local modelling} that interprets spectral shaping as curvature-dependent reweighting of the update along \emph{local curvature directions} of the loss landscape. 
Along these directions, it jointly tracks the \emph{residual signal}, i.e., the remaining parameter distance to a nearby local optimum, and the stochastic gradient noise introduced by minibatch sampling. 
This mode-wise decomposition reveals an interesting signal--noise trade-off: decreasing $p$ increases residual-signal contraction in ``flat'', lower-curvature directions, but also amplifies noise along these same directions.
It further suggests that as training progresses, the remaining residual signal becomes less concentrated in high-curvature directions and relatively more prominent in small, flat curvature directions~(\Cref{sec:modeling}).

This leads to a stage-dependent finding: positive $p$ helps early by emphasizing high-curvature directions and accelerating residual-signal contraction, whereas a mildly negative $p$ helps later by refocusing the updates towards flat directions that still carry useful residual signal.
Thus, we uncover a previously overlooked late-stage training behavior: dynamically shifting emphasis toward flat directions further improves optimization, a stage-dependent advantage that fixed transformations such as Muon cannot capture.
Our empirical observations support the predictions of our modelling and show that adapting $p$ across training improves performance~(\Cref{sec:positive,sec:negative,sec:validation}).

Motivated by these observations, we propose \sys, a dynamic spectral shaping algorithm that adapts the spectral exponent $p$ over training~(\Cref{sec:method}).
It leverages a simple decreasing logistic schedule for $p$, interpolating from positive values early in training to mildly negative values later.
To realize our scheduled spectral shaping efficiently, \sys extends  Newton--Schulz approximation to approximate $U\Sigma^pV^\top$ for varying values of $p$, avoiding full SVD and retaining the per-step cost as Muon.
Extensive experiments across model sizes and architectures show that \sys consistently achieves lower validation loss than Muon across training settings, while reaching target losses with up to \textbf{26.5\%} fewer training steps. To summarize, our main contributions are:
\begin{enumerate}[leftmargin=*, noitemsep]
    \item We introduce a dynamic spectral-shaping perspective for matrix-valued updates, reframing Muon-style optimization from a fixed spectral operation into the adaptive problem of choosing a suitable spectral exponent $p$ as training dynamics evolve.
    \item We develop a noise-aware local model (\cref{sec:obs}) that explains how the right choice for the spectral exponent changes across training stages through a trade-off between residual-signal reduction and stochastic-noise amplification across local curvature directions.
    \item Guided by this model, we uncover a surprising stage-dependent regime: positive spectral exponents help early training, whereas (previously overlooked) negative exponents improve late-stage optimization by emphasizing flat directions that retain useful residual signal.
    \item In \Cref{sec:method}, we propose \sys, a simple and efficient dynamic spectral-shaping algorithm that adapts the spectral exponent throughout training, yielding consistent improvements over Muon across training settings (\Cref{sec:evaluation}). 
\end{enumerate}

\subsection{Other Related Work}\label{sec:prelim}

\noindent\textbf{Muon and Spectral Shaping.}
Recent works have shown that orthonormalizing matrix-shaped momentum can substantially improve neural network / LLM training~\citep{jordan2024muon,liu2025muonscalablellmtraining}. 
Muon uses Newton--Schulz iterations to produce orthonormalized updates for hidden layers, and recent works explain its effectiveness through trust-region, norm-constrained, and spectral preconditioning perspectives~\citep{kovalev2025understandinggradientorthogonalizationdeep,pethick2025training,chen2026muon,ma2026preconditioningbenefitsspectralorthogonalization}. 
A growing line of work studies variants of matrix updates, including specific fixed positive choices of $p$ within the $U\Sigma^pV^\top$, as well as spike-aware, anisotropic, or mode-guided transformations~\citep{qi2026delvingmuonbeyonddeep,huang2026spectrarethinkingoptimizersllms,yang2026prismstructuredoptimizationanisotropic,lu2026muonspectralguidanceefficient}. 
More broadly, matrix-aware optimizers such as Shampoo, SOAP, and PolarGrad exploit non-diagonal geometry and spectral structure beyond coordinate-wise scaling~\citep{pmlr-v80-gupta18a,shi2023distributeddataparallelpytorchimplementation,vyas2025soap,lau2026polargradclassmatrixgradientoptimizers}. 
These methods establish spectral structure as an important design axis, but they largely use fixed or task-specific transformations.
In contrast, this paper studies whether the preferred spectral bias should evolve across training stages, systematically developing a dynamic spectral shaping rule.

\noindent\textbf{Dynamic Scheduling in Optimization.}
Dynamic schedules are widely used to adapt optimization behavior across training phases, including learning-rate annealing, restarts, warmup, cyclical policies, iteration-dependent scaling, and studies of stage-dependent subspace behavior~\citep{loshchilov2017sgdr,smith2018disciplinedapproachneuralnetwork,smith2018superconvergencefasttrainingneural,Ma_Yarats_2021,You2020Large,pmlr-v291-gao25a,deng2026suspiciousalignmentsgdfinegrained}. 
Unlike schedules that mainly modulate scalar hyperparameters, \sys dynamically schedules the spectral shape of matrix-valued updates, allowing the emphasis  we put over different spectral directions to evolve across training stages.
See extended related work in~\Cref{app:related}.

\section{Motivating Dynamic Spectral Shaping with an Idealized Model}\label{sec:obs}
Although Muon-style methods~\citep{jordan2024muon,liu2025muonscalablellmtraining} have shown strong empirical gains from the orthonormalized update, recent analyses suggest that their effectiveness can vary across training conditions~\citep{qi2026delvingmuonbeyonddeep,2026useusemuonsimplicity}.
This raises the question of whether fixed Muon-style spectral shaping remains optimal throughout training.
Motivated by this, we systematically vary \textbf{\emph{spectral component}} $p$ in the spectral-shaping family $U\Sigma^pV^\top$ and study its effect on optimization.
Before proposing our method, we first develop a simple noise-aware local model, using a standard local quadratic and gradient-noise approximation, to isolate how $p$ controls the trade-off between useful training signal and gradient noise~(\Cref{sec:modeling}). We note that our modelling and analysis make several simplifying approximations (which we empirically validate).
Our goal is not to provide a full end-to-end convergence theory, but rather to introduce a minimal set of reasonable approximations that yield mechanistic insights into the role of the spectral exponent $p$ in training dynamics. 
In particular, our modelling predicts two stage-dependent regimes: positive $p$ can benefit early training, whereas mildly negative $p$ can improve late-stage training.
We \emph{empirically validate these predictions} in~\Cref{sec:positive,sec:negative,sec:validation}, showing that our simplified modelling can serve as a useful and \emph{predictive} guide for designing dynamic spectral shaping.

\subsection{A Noise-Aware Local Model for Spectral Shaping}\label{sec:modeling}
Our starting point is to consider one such weight matrix $W_t\in\mathbb{R}^{m\times n}$ at training step $t$, and take its stochastic gradient $G_t = U_t \Sigma_t V_t^\top$ as the update matrix.
Applying spectral shaping with exponent $p$ gives
$D_t^{(p)} := U_t \Sigma_t^p V_t^\top$.
With the learning rate $\eta$, the parameter update is
\vskip -0.16in
\begin{equation}
    W_{t+1}= W_t-\eta D_t^{(p)} = W_t - \eta{{(G_t G_t^\top)^{\frac{p-1}{2}}}} G_t. 
\label{eq:update}
\end{equation}
\vskip -0.05in

Let $L(W)$ denote the population loss, and let $W^\star$ be a nearby local minimizer with $\nabla L(W^\star)\approx 0$.
We define the \emph{\textbf{residual signal}} as $E_t:=W_t-W^\star$, measuring the remaining error relative to $W^\star$.
To analyze how the shaped update in~\Cref{eq:update} affects optimization, we study the evolution of this residual signal under the update.
Since the update is driven by the gradient, we relate the gradient to the residual signal $E_t$ by performing a locally linear approximation of the population gradient around $W^\star$~\citep{nocedal2006numerical}.
We further use a one-sided Kronecker-factored approximation to the local Hessian action, in the spirit of K-FAC~\citep{pmlr-v37-martens15}.
We apply this approximation over a short training window in which the effective local curvature is empirically stable (see Figure~\ref{fig:grad2-curv-alignment} for empirical support):
\begin{equation}
\nabla L(W_t) \approx \nabla L(W^\star)+\nabla^2 L(W^\star)[E_t] \approx \kappa_t H E_t,
\label{eq:local_approx}
\end{equation}
\vskip -0.1in
where $H=Q\Lambda Q^\top$ is a normalized effective local curvature matrix with $\Lambda=\mathrm{diag}(h_1,\dots,h_m)$, $h_i\in(0,1]$, and $\max_i h_i =1 $.
The scalar $\kappa_t>0$ captures the overall curvature scale.
The eigenvectors of $H$ define the \emph{modes} of the local loss landscape, where each mode corresponds to one curvature direction and $h_i$ measures the curvature along that direction.
Modes with larger $h_i$ are called \emph{\textbf{strong modes}} and modes with smaller $h_i$ are called \emph{\textbf{flat modes}}. 
To account for stochasticity in the actual training update, we decompose the stochastic gradient $G_t$ into the population gradient and a zero-mean noise $\Xi_t$, following standard unbiased SGD assumptions~\citep{doi:10.1137/16M1080173}. 
Plugging into (\ref{eq:local_approx}) gives: 
\vskip -0.17in
\begin{equation}
G_t = \nabla L(W_t) + \Xi_t \approx \kappa_t H E_t + \Xi_t.
\label{eq:stochastic_grad_matrix_model}
\end{equation}
\vskip -0.05in
Since the shaped update depends on $(G_tG_t^\top)^{\frac{p-1}{2}}$, we relate the spectral structure of $G_tG_t^\top$ to the effective local curvature.
Motivated by the common use of gradient second moments as Fisher-type proxies for local curvature~\citep{NEURIPS2019_46a558d9,JMLR:v21:17-678}, we use a curvature-aligned surrogate:
$(G_t G_t^\top)^{\frac{p-1}{2}} \approx \alpha_t^{\frac{p-1}{2}} H^{\frac{p-1}{2}}$,
where $\alpha_t> 0$ is a scalar factor. 
This approximation models spectral shaping as a curvature-dependent reweighting.
The underlying gradient-curvature alignment is empirically supported in Appendix~\Cref{fig:grad2-curv-alignment}.
Substituting this approximation into~\Cref{eq:update}, absorbing $\alpha_t^{\frac{p-1}{2}}$ into the effective learning rate $\eta_t$, and using~\Cref{eq:stochastic_grad_matrix_model}, the residual signal evolves as
\vskip -0.16in
\begin{equation}
E_{t+1} = E_t-\eta_t H^{\frac{p-1}{2}}(\kappa_t HE_t+\Xi_t) = \left(I-\eta_t \kappa_t H^{\frac{p+1}{2}}\right)E_t - \eta_t H^{\frac{p-1}{2}}\Xi_t.
\label{eq:matrix_error_dynamics_model}
\end{equation}
\vskip -0.06in

\noindent\textbf{Mode-Wise Signal--Noise Tradeoff.}
Based on the intuitive ``idealized'' analysis above, we now arrive at the main training dynamics equation that we focus on. 
Since $p$ acts through powers of the curvature matrix $H=Q\Lambda Q^\top$ in~\Cref{eq:matrix_error_dynamics_model}, its effect can vary across curvature directions.
This motivates a mode-wise analysis, where we project the residual signal $E_t$ and noise $\Xi_t$ onto the eigenbasis of $H$.
Define $\widetilde{E}_t := Q^\top E_t$ and $Z_t := Q^\top \Xi_t$.
Let $\delta_{i,t}$ and $\xi_{i,t}$ denote the $i$-th coordinates of $\widetilde{E}_t$ and $Z_t$.
Then mode $i$ evolves as
\begin{equation}
\boxed{\delta_{i,t+1} = \left(1-\eta_t \kappa_t h_{i}^{\frac{p+1}{2}}\right)\delta_{i,t} - \eta_t h_{i}^{\frac{p-1}{2}}\xi_{i,t}.}
\label{eq:scalar_dynamics_model}
\end{equation}

We next consider the squared residual for each mode.
Assume that the mode-wise noise is conditionally zero-mean with
$\mathbb E[\xi_{i,t}\mid \delta_{i,t}] = 0$ and
$\mathbb E[\xi_{i,t}^2\mid \delta_{i,t}] = c_{i,t}$,
where $c_{i,t}$ denotes the noise level of mode $i$ at step $t$.
Then~\Cref{eq:scalar_dynamics_model} gives
\begin{equation}
\boxed{\mathbb E[\delta_{i,t+1}^2 \mid \delta_{i,t}] = \left(1-\eta_t \kappa_t h_i^{\frac{p+1}{2}}\right)^2 \delta_{i,t}^2 + \eta_t^2 h_i^{p-1}c_{i,t}.}
\label{eq:one_step_second_moment}
\end{equation}
Thus, $p$ induces a \emph{mode-wise signal--noise trade-off}.
The deterministic multiplier $1-\eta_t\kappa_t h_i^{(p+1)/2}$ controls residual-signal contraction: within the stable range $(0,1)$, larger $h_i^{(p+1)/2}$ makes this multiplier smaller and contracts $\delta_{i,t}$ faster.
Increasing $p$ therefore favors contraction in strong, high-curvature modes, whereas decreasing $p$ increases the relative contraction strength in flat modes.
However, the stochastic term is scaled by $h_i^{p-1}$, so decreasing $p$ also amplifies noise most strongly in flat modes.
Without noise, $p=-1$ would maximize contraction in flat modes; with noise, the choice of $p$ must balance residual-signal contraction against noise amplification.
This decomposition also provides the basis for the mode-wise predictions tested in~\Cref{sec:validation}, where we empirically estimate the residual-signal ``energy'' $\delta_{i,t}^2$ and noise level $c_{i,t}$ during training.

\begin{tcolorbox}[left=0mm, right=0mm, top=0mm, bottom=0mm, boxrule=0.5pt, breakable]
\textbf{Takeaway:}
The spectral exponent $p$ controls a mode-wise signal--noise tradeoff.
Larger $p$ accelerates residual signal contraction in strong modes, while smaller $p$ shifts more emphasis toward flat modes but also amplifies noise along them.
\end{tcolorbox}

\subsection{Why a Slightly Negative Spectral Exponent Is Preferred in the Late Stage}\label{sec:negative}
Building on the mode-wise signal--noise tradeoff in~\Cref{eq:one_step_second_moment}, we assess how different spectral exponents affect training performance.
We focus on $p \in [-1,1]$, covering the representative cases in~\Cref{sec:prelim}.
At a high level, training improves when the update reduces residual signal in modes where substantial signal remains.
Since varying $p$ changes which modes are emphasized, the preferred choice of $p$ should depend on how the residual signal is distributed across modes during training.

\noindent\textbf{Why Residual Signal Concentrates in Flat Modes Late in Training.}
For $h_i\in(0,1]$, the contraction strength $h_i^{\frac{p+1}{2}}$ increases with $h_i$.
Thus, the residual signal in strong modes tends to decay earlier, whereas the signal in flat modes decays more slowly and can remain substantial later in training.
This suggests that the residual signal becomes relatively more concentrated in flat modes as training progresses.
This is also consistent with the local gradient approximation in~\Cref{eq:local_approx}.
Projecting the population gradient onto mode $i$ gives $g_{i,t} \approx \kappa_t h_i \delta_{i,t}$, hence $\delta_{i,t} \approx \frac{g_{i,t}}{\kappa_t h_i}$.
Thus, for comparable projected gradient magnitudes, a smaller curvature $h_i$ corresponds to a larger residual signal.

\noindent\textbf{When a Mildly Negative Exponent Helps in the Late Stage.}
Once the residual signal becomes more concentrated in flat modes in the late training stage, it can be beneficial to place relatively more emphasis on those modes rather than on strong modes whose residual signal has already decayed.
Decreasing $p$ achieves this effect by allocating relatively more contraction to flat modes.
However, from~\Cref{eq:one_step_second_moment}, lowering $p$ also increases the noise level, especially in flat modes.
Thus, a negative $p$ can improve over $p=0$ only when the residual signal in flat modes is large enough relative to the noise.
This also explains why the exponent should be only mildly negative: otherwise, noise amplification can outweigh the benefit from reducing the residual signal and degrade optimization.

\begin{figure*}[t]
    \centering
    \includegraphics[width=0.85\linewidth]{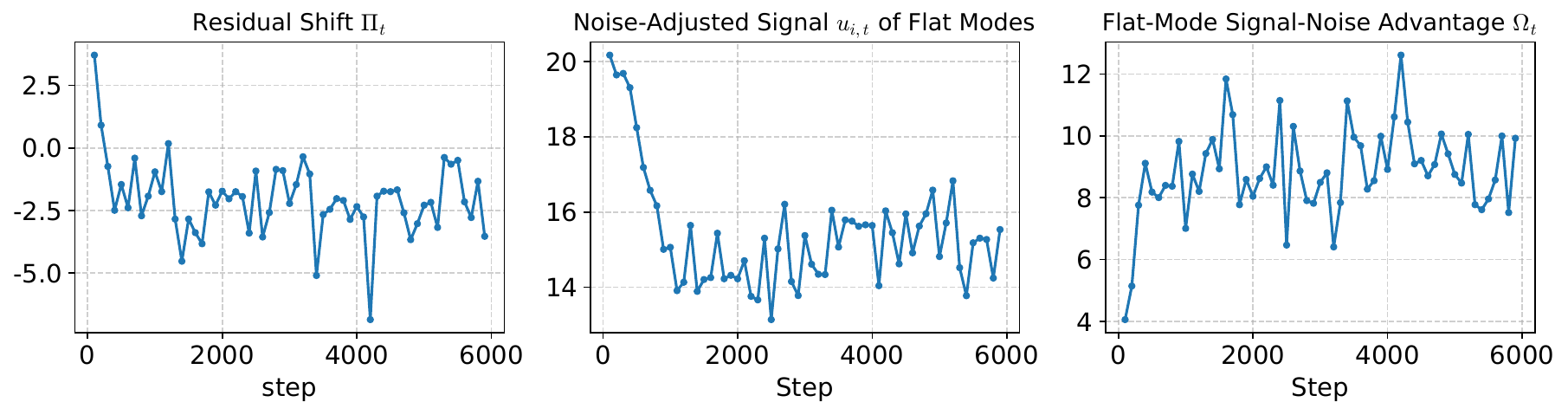}
    \caption{
        Validation of the mode-wise model predictions.
        \textbf{Left:} The residual-shift metric $\Pi_t$ decreases during training and becomes negative, indicating that the residual signal shifts from strong, high-curvature modes toward flat modes.
        \textbf{Middle:} The noise-adjusted signal in flat modes remains substantially positive, indicating that flat modes retain the residual signal that is large relative to their noise level.
        \textbf{Right:} The flat-mode signal-noise advantage remains positive, showing that flat modes have a favorable signal-noise tradeoff relative to strong modes in the late stage.
    }
    \label{fig:signal}
    \vskip -0.14in
\end{figure*}

\begin{figure*}[t]
    \centering
    \includegraphics[width=0.85\linewidth]{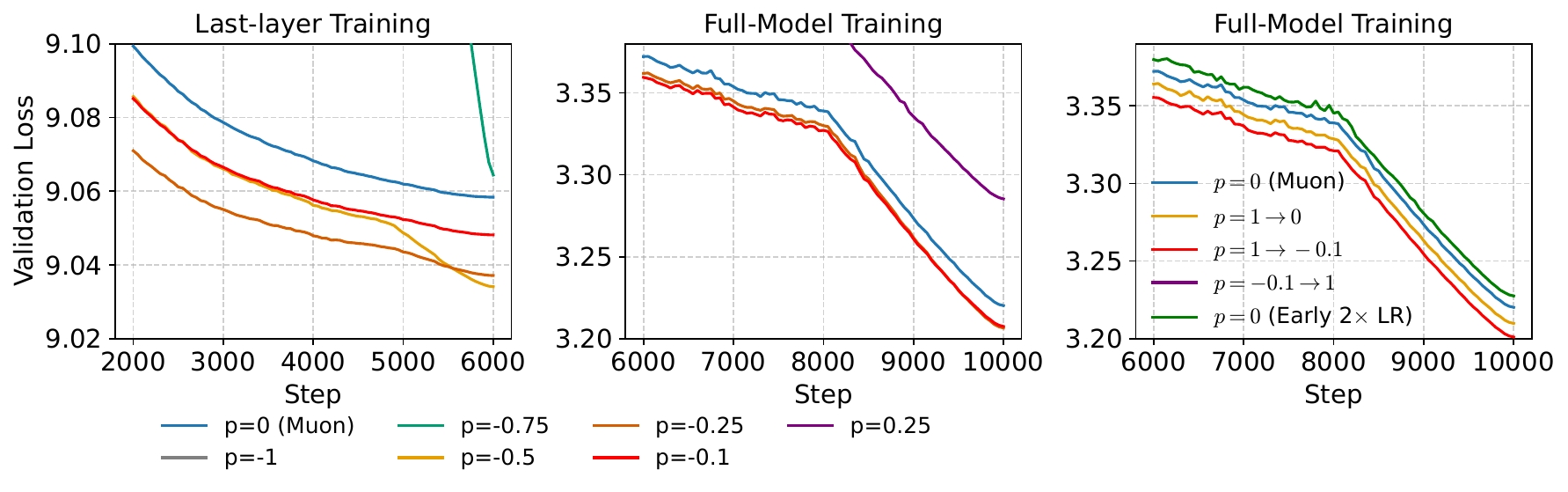}
    \caption{
        Training performance of stage-dependent spectral shaping.
        \textbf{Left/Middle:} Mildly negative exponents improve late-stage validation loss in both last-layer and full-model training, while overly negative and late-positive exponents degrade performance.
        \textbf{Right:} Early positive $p$ improves full-model training, with the positive-to-negative schedule achieving the lowest validation loss. 
        In contrast, simply doubling the early learning rate or reversing the schedule from negative to positive performs worse than Muon.
        The panels use capped y-axes; uncapped versions are shown in Appendix~\Cref{fig:train_obs_full}.
    }

    \label{fig:train_obs}
    \vskip -0.232in
\end{figure*}

\subsection{Validating Predictions of Our Model}\label{sec:validation}
We empirically examine whether the signal-noise conditions predicted by our model arise under Muon ($p=0$).
Using a GPT-style model with hidden dimension 768, we freeze all parameters except one target matrix in the final transformer block and estimate its mode-wise curvature, residual signal, and noise.
Full setup and estimation details are provided in~\Cref{app:obs_exp}.

\emph{\underline{Empirical Modes and Proxies.}}
In our analysis, modes are curvature directions, which are expensive to track directly in practice.
We therefore use the singular directions of the Muon update as empirical modes, since spectral shaping reweights singular values along these directions.
For each empirical mode $i$ at step $t$, we estimate local curvature $\hat{h}_{i,t}$, mode-wise noise level $\hat{c}_{i,t}$, and residual-signal energy $\hat{\delta}_{i,t}^{2}$.
Specifically, $\hat{h}_{i,t}$ is estimated via Hessian-vector products, $\hat{c}_{i,t}$ from the variance of independent mini-batch gradient projections, and $\hat{\delta}_{i,t}^{2}$ from fixed-probe gradient projections using the local gradient approximation in~\Cref{eq:local_approx}.

\emph{\underline{Validating Residual-Signal Concentration in Flat Modes.}}
To test whether the residual signal concentrates in flat modes, we sort modes by curvature $\hat{h}_{i,t}$ and define $\mathcal{S}$ and $\mathcal{F}$ as the top-8 highest-curvature and bottom-8 lowest-curvature modes, respectively.
We summarize each bucket by the median $\log$ residual signal energy and define the \emph{residual shift} as 
$ \Pi_t = \operatorname{med}_{i\in\mathcal{S}} \log \hat{\delta}_{i,t}^{2} - \operatorname{med}_{i\in\mathcal{F}} \log \hat{\delta}_{i,t}^{2}.$
Thus, $\Pi_t<0$ indicates that flat modes have a larger residual signal than strong modes.
Figure~\ref{fig:signal} (left) shows that $\Pi_t$ steadily decreases and becomes mostly negative after roughly $500$ steps, supporting the predicted late-stage concentration of residual signal in flat modes.

\emph{\underline{Validating Flat-Mode Signal After Accounting for Noise.}}
Since a negative exponent also amplifies noise, we measure whether the flat-mode residual signal remains useful after accounting for noise.
Therefore, we define the mode-wise \emph{noise-adjusted signal}
$u_{i,t}:=\log \hat{\delta}_{i,t}^{2} - \log \hat{c}_{i,t}$,
which measures the residual signal relative to the noise, and the \emph{flat-mode signal-noise advantage}
$\Omega_t = \operatorname{med}_{i\in\mathcal F}u_{i,t} - \operatorname{med}_{i\in\mathcal S}u_{i,t}$,
which compares this noise-adjusted signal between flat and strong modes.
Figures~\ref{fig:signal} (middle, right) show that the noise-adjusted signal $u_{i,t}$ in flat modes remains substantially positive and that $\Omega_t$ quickly increases and remains positive.
This indicates that flat modes retain the residual signal large enough relative to noise, matching the condition under which a mildly negative exponent can improve over the Muon choice $p=0$.
We provide additional empirical analysis in~\Cref{app:noise}.

\noindent\textbf{Training Performance for Mildly Negative $p$.}
We next test whether the mode-wise analysis translates into improved training performance by switching from $p=0$ to a negative exponent at step 500, when the residual signal begins to concentrate in flat modes.
We first compare negative exponents $p \in \{-0.1,-0.25,-0.5,-0.75,-1\}$ in both last-layer-only and full-model training.
As shown in Figure~\ref{fig:train_obs} (left, middle), a mildly negative $p$ provides the best performance: $p=-0.1$ and $p=-0.25$ outperform the Muon default $p=0$, while more aggressive choices such as $p=-0.75$ and $p=-1$ become unstable and perform much worse.
This agrees with our analysis: moderate emphasis on flat modes helps late-stage optimization, but overly negative $p$ amplifies noise and degrades performance.
Conversely, switching to a positive exponent ($p=0.25$) at the same step degrades full-model training performance, as shown in Figure~\ref{fig:train_obs} (middle). 
This result further supports our prediction that late-stage gains specifically stem from emphasizing flat modes.

\begin{tcolorbox}[left=0mm, right=0mm, top=0mm, bottom=0mm, boxrule=0.5pt, breakable]
\textbf{Takeaway:}
In late training, residual signal shifts toward flat modes, making a slightly negative spectral exponent preferable because it allocates more contraction to these directions.
However, this preference is mild: if $p$ is too negative, noise amplification can outweigh the contraction benefit and hurt training.
This matches our empirical results, where mildly negative values improve performance while more aggressive negative values degrade it.
\end{tcolorbox}

\subsection{Why a Positive Spectral Exponent Can Help in the Early Stage}
\label{sec:positive}
Beyond the late-stage preference for a mildly negative exponent, we ask whether the Muon default $p=0$ is also optimal early in training. 
Our analysis suggests otherwise: early residual signal can remain concentrated in high-curvature modes, and increasing $p$ accelerates contraction of these strong-mode residual signal while reducing the noise-amplification factor $h_i^{\frac{p-1}{2}}$.
Thus, before the residual distribution shifts toward flat modes, a positive exponent can accelerate early optimization by prioritizing the strong-mode residual signal with limited noise amplification.

\noindent\textbf{Training Performance with an Early Positive Exponent.}
To test this prediction, we use a simple two-stage schedule that sets $p=1$ for the first 500 steps, then switches to either $p=0$ or $p=-0.1$.
As Figure~\ref{fig:train_obs} (right) shows, both early-positive schedules achieve lower validation loss than the fixed Muon baseline, supporting the view that prioritizing strong modes early improves the optimization trajectory.
Notably, the schedule transitioning from $p=1$ to $p=-0.1$ yields the best performance.
This validates our stage-dependent picture: a positive exponent accelerates early training, while a mildly negative exponent improves late-stage optimization as the residual signal becomes relatively more concentrated in flat modes.
We further test two alternative explanations.
First, to rule out the possibility that the early positive $p$ merely acts as a larger effective learning rate, we test a Muon variant that doubles the learning rate during the first 500 steps while keeping $p=0$.
This variant performs worse than fixed Muon, indicating that the gain from an early positive $p$ is distinct from simple step-size scaling.
Second, we evaluate a reverse schedule that switches from $p=-0.1$ to $p=1$ at step 500.
This schedule also performs worse than fixed Muon, consistent with our analysis that positive $p$ should be applied early, while mildly negative $p$ is beneficial only later.

\begin{tcolorbox}[left=0mm, right=0mm, top=0mm, bottom=0mm, boxrule=0.5pt, breakable]
\textbf{Takeaway:}
A positive early-stage exponent helps reduce strong-mode residual signal before the remaining residual signal shifts toward flat modes, yielding lower validation loss than fixed Muon.
\vspace{-.5mm}
\end{tcolorbox}

\section{\sys: Dynamic Spectral Shaping}\label{sec:method}

The above analysis and observations motivate \sys~(\Cref{alg:dyn-full} in~\Cref{app:algo}), which dynamically adapts spectral shaping by {monotonically} decreasing the spectral exponent from a positive early-stage value to a mildly negative late-stage value, while maintaining computational efficiency.

\noindent\textbf{Logistic Scheduling of the Spectral Exponent.}
Although the residual-signal distribution across modes can guide the choice of $p$, estimating it online would require additional forward and backward passes.
We therefore use a simple logistic schedule to approximate a smooth decreasing transition of $p_t$ over training without this extra cost.
Given the current training step $t$ and total steps $T$, we set
\begin{equation*}
u_t=\frac{t/T-\tau}{w},
\qquad
a_t=\frac{1}{1+\exp(u_t)},
\qquad
p_t=p_{\min}+a_t(p_{\max}-p_{\min}).
\end{equation*}
Here, $\tau$ controls the transition point and $w$ controls the transition width, with a smaller $w$ producing a sharper switch.
We set $p_{\max}=1$ and $p_{\min}=-0.25$, where $p_{\min}=-0.25$ is the best-performing negative exponent based on our observations in~\Cref{sec:negative} (we also ablate $p_{\min}$; see Section \ref{sec:evaluation}).

\noindent\textbf{Efficient Updates.} 
Given a scheduled exponent $p_t$, exact SVD can realize the corresponding fractional spectral shaping, but it is computationally expensive.
Muon avoids SVD by approximating the $p=0$ polar update with a small fixed number of \tsf{Newton--Schulz} (NS) iterations, but it does not directly support arbitrary exponents.
To obtain an efficient implementation, \sys uses an equivalent factorization of the target spectral shaping~(Appendix~\Cref{alg:efficient}).
For the normalized input $X_n=X/\|X\|_F$ and $A:=X_nX_n^\top$, we first write
\vskip -0.15in
\begin{equation}
\label{eq:factor}
U\Sigma^pV^\top=(X_nX_n^\top)^{\frac{p-1}{2}}X_n
=\textcolor{red}{(X_nX_n^\top)^{\frac p2}} \textcolor{blue}{(X_nX_n^\top)^{-\frac12}X_n} = \textcolor{red}{A^{\frac p2}} \textcolor{blue}{A^{-\frac12}X_n}.
\end{equation}
\vskip -0.1in
Since $Y_\mu = \textcolor{blue}{A^{-\frac12}X_n}$ corresponds to the Muon update, it can be efficiently approximated by NS. 
It remains to approximate the left correction $\textcolor{red}{A^{\frac p2}}$. 
When this factor acts as a mild spectral correction, it introduces only a small adjustment on top of the Muon update.
We thus approximate it by a second-order Taylor expansion around the identity $I$.
Letting $E=A-I, \delta=p/2$, we have
$
\textcolor{red}{A^{\frac p2}} \approx C = I + \delta E + \frac{1}{2}\delta(\delta-1)E^2.
$
The target spectral shaping is then given by 
$
\widetilde{X} = \|X\|_F^p C Y_\mu.
$ For $X\in\mathbb{R}^{m\times n}$ with $m\le n$, \sys adds only one polynomial correction of cost $O(m^2n+m^3)$ on top of NS computations.
Thus, our method has the \emph{same asymptotic complexity} as NS-based Muon. Figure \ref{fig:data_pmin} (right) shows that our approximation closely tracks exact SVD throughout training.

\noindent\textbf{Stable Anchoring for Positive Exponents.}
\sys implements the spectral shaping scheduled exponent $p_t$ through a simple stage-wise scheme. 
The main reason is stability: when $p_t$ is positive and sufficiently large, the correction factor $\textcolor{red}{A^{\frac p2}}$ is no longer a mild adjustment to the Muon update, so the Taylor approximation around $A\approx I$ can become unreliable.
To avoid this instability, \sys anchors the positive regime to two stable operators (lines 7--10 in~\Cref{alg:dyn-full}).
\sys uses the original update when $p_t \geq 0.25$.
For $p_t \in [0, 0.25)$, \sys applies standard NS orthogonalization, recovering the Muon-style update.
For $p_t \in [p_{\min}, 0)$, the exponent is only mildly negative, so \sys uses the efficient spectral approximation described above for a continuous schedule.

\section{Evaluation}\label{sec:evaluation}

\noindent\textbf{Models and Datasets.}
We evaluate \sys on two decoder-only Transformer families: GPT-style models at multiple scales following modded-nanoGPT~\cite{modded_nanogpt_2024} and a Qwen-style model, with detailed configurations summarized in~\Cref{tab:gpt_scales,tab:qwen_scales}.
The GPT-style models use rotary position embeddings~\cite{su2024roformer}, RMSNorm, and squared ReLU MLPs~\cite{NEURIPS2021_2f3c6a4c}.
The Qwen-style model uses pre-normalized Transformer blocks with RMSNorm, grouped-query attention, and gated SiLU MLPs.
All models use sequence length $1024$ and global batch size $512$.
Our main experiments use 10B tokens from FineWeb, and we additionally evaluate on FineWeb-Edu~\cite{NEURIPS2024_370df50c}.
To study training-budget scaling, we vary the number of training tokens from 2.5B to 20B.

\noindent\textbf{Baselines.}
We compare \sys against Muon~\cite{jordan2024muon} and AdamW~\cite{loshchilov2018decoupled}.
Muon is our primary and most directly relevant baseline, while AdamW serves as a standard, widely used optimizer baseline.
We further include NorMuon~\citep{li2025normuonmakingmuonefficient} as an additional Muon-variant baseline in~\Cref{app:results}.
Unless otherwise specified, the default learning rates are $0.01$ for Muon and \sys, and $0.002$ for AdamW.
We also vary learning rates for all methods.
By default, \sys uses $p_{\max}=1$ and $p_{\min}=-0.25$. 
Additional evaluation setup details are provided in~\Cref{app:exp_set}.

\begin{figure*}[t]
    \centering
    \includegraphics[width=0.9\linewidth]{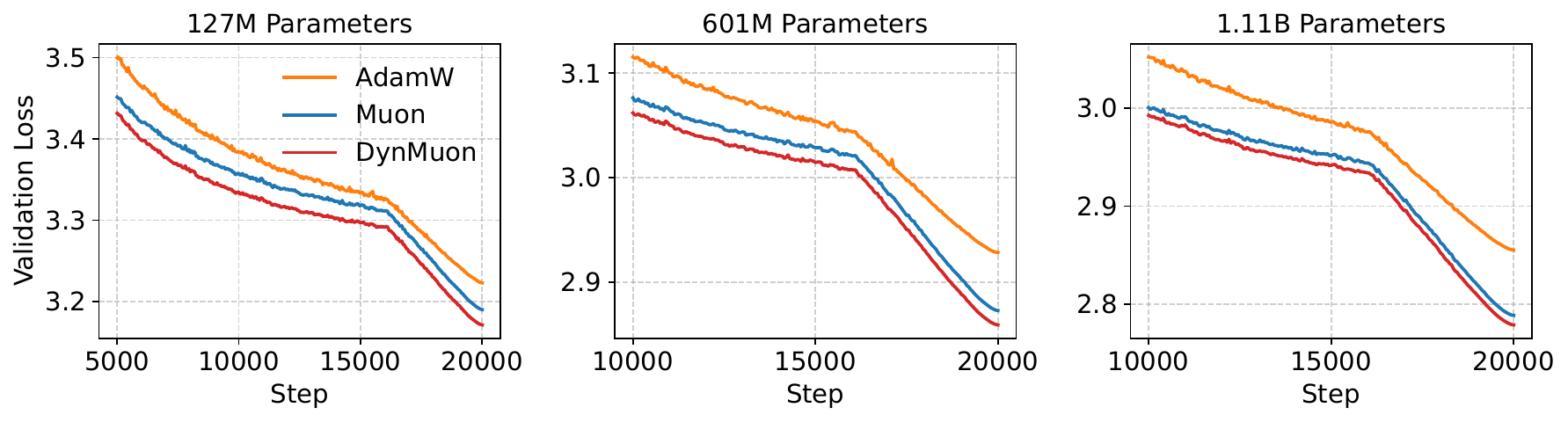}
    \caption{
    Validation loss trajectories across three model scales trained on 10B tokens. 
    \sys consistently achieves the lowest validation loss across all three model scales.
    }
    \label{fig:main}
    \vskip -0.2in
\end{figure*}


\begin{table}[t]
\centering
\small
\caption{
Performance and efficiency of \sys relative to Muon across GPT-style model scales.
Steps to Target uses the validation loss reached by Muon at 80\% of training as the target.
Step Saving reports the relative step reduction, and Per-Step Time is the average ms/step. 
}
\setlength{\tabcolsep}{4pt}
\label{tab:efficiency}
\scalebox{0.85}{
\begin{tabular}{llcccc}
\toprule
\textbf{Tokens} &
\textbf{Method (Size)}
& \textbf{Best Val. Loss ($\downarrow$)}
& \textbf{Steps to Target ($\downarrow$)}
& \textbf{Step Saving ($\uparrow$)}
& \textbf{Per-Step Time (ms)} \\
\midrule
\multirow{6}{*}{10B}
& Muon (127M) 
& 3.190 & 16000 & 0.0\% & 1142.4  \\
& \sys (127M)
& \textbf{3.171} & \textbf{12500} & \textbf{21.9\%} & 1150.3  \\
\cmidrule(lr){2-6}
& Muon (601M)
& 2.872 & 16000 & 0.0\% & 4121.7 \\
& \sys (601M)
& \textbf{2.858} & \textbf{13950} & \textbf{12.8\%} & 4200.1  \\
\cmidrule(lr){2-6}
& Muon (1.1B)
& 2.788 & 16000 & 0.0\% & 6883.3  \\
& \sys (1.1B)
& \textbf{2.776} & \textbf{14300} & \textbf{10.6\%} & 7055.8 \\
\midrule
\multirow{6}{*}{20B}
& Muon (127M)
& 3.139 & 30400 & 0.0\% & 1137.3 \\
& \sys (127M)
& \textbf{3.124} & \textbf{22350} & \textbf{26.5\%} & 1151.8 \\
\cmidrule(lr){2-6}
& Muon (601M)
& 2.808 & 30400 & 0.0\% & 4126.2 \\
& \sys (601M)
& \textbf{2.797} & \textbf{25000} & \textbf{17.8\%} & 4184.8 \\
\cmidrule(lr){2-6}
& Muon (1.1B)
& 2.722 & 30400 & 0.0\% & 6889.77 \\
& \sys (1.1B)
& \textbf{2.713} & \textbf{26450} & \textbf{13.0\%} & 6910.1 \\
\bottomrule
\end{tabular}
}
\vskip -0.05in
\end{table}


\begin{figure*}[t]
    \centering
    \includegraphics[width=0.85\linewidth]{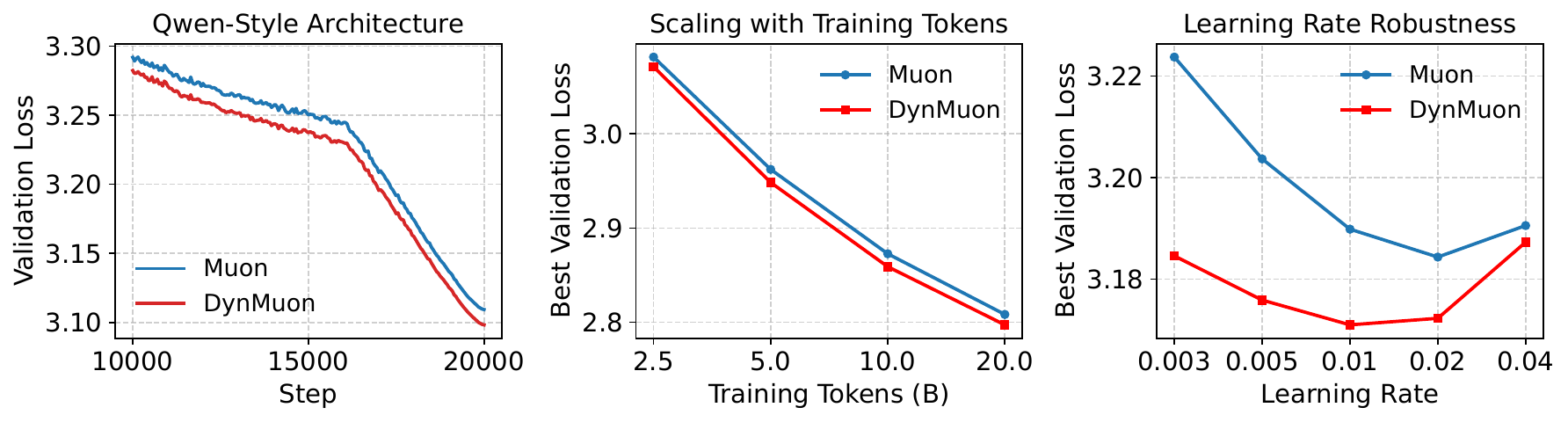}
    \caption{
        \sys outperforms Muon over architectures, training-token budgets, and learning rates. 
    }
    \label{fig:token_lr}
    \vskip -0.25in
\end{figure*}

\begin{figure*}[t]
    \centering
    \includegraphics[width=0.9\linewidth]{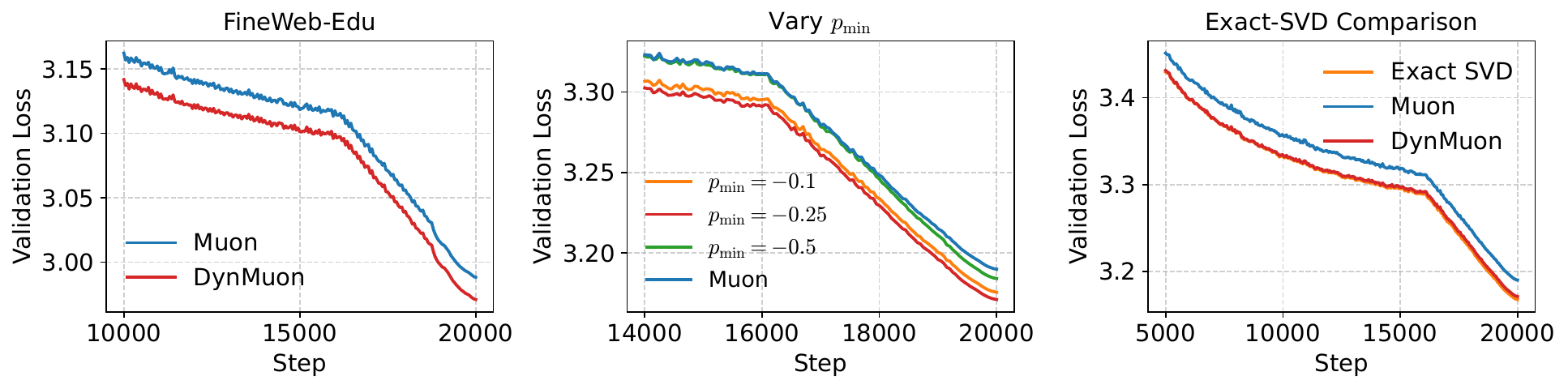}
    \caption{
        Additional experiments for \sys across corpora, $p_{\min}$ choices, and spectral-shaping implementations.
        \textbf{Left:} \sys outperforms Muon on FineWeb-Edu.
        \textbf{Middle:} mildly negative $p_{\min}$ values perform best.
        \textbf{Right:} our spectral shaping approximations closely tracks exact SVD.
    }
    \label{fig:data_pmin}
    \vskip -0.13in
\end{figure*}

\begin{figure*}[t]
    \centering
    \includegraphics[width=0.9\linewidth]{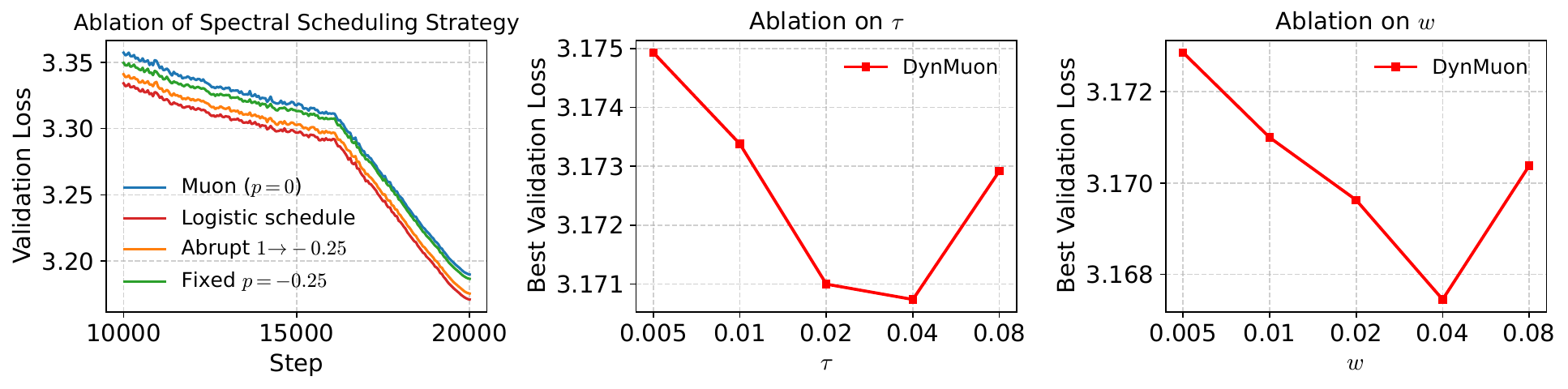}
    \caption{
       Ablation of spectral scheduling strategies and logistic schedule parameters $(\tau,w)$.
    }
    \label{fig:ablation}
    \vskip -0.24in
\end{figure*}

\noindent\textbf{Main Results.}
We train GPT-style models at three scales on FineWeb using both 10B and 20B token budgets.
As shown in Figure~\ref{fig:main}, \sys consistently achieves the lowest validation loss across all three model scales compared with baselines.
The improvement over Muon is clear in the late training stage, where scheduling the spectral exponent toward mildly negative values provides a stable advantage.
Table~\ref{tab:efficiency} quantifies the practical significance of these gains in terms of both step efficiency and runtime overhead.
For each model size and token budget, we define a fixed target as the validation loss reached by Muon at 80\% of training, and record the first step at which \sys reaches it.
Across model scales, \sys reaches the target \textbf{10.6--26.5\%} earlier than Muon, requiring substantially fewer training steps to reach the same validation loss.
Meanwhile, \sys has a per-step time ratio of only \textbf{1.003--1.025$\times$} relative to Muon, indicating negligible additional runtime cost.
Thus, \sys improves final performance and step efficiency with minimal per-step overhead.


\noindent\textbf{Model Architecture.}
To evaluate transferability beyond GPT-style models, we train a 171M Qwen-style decoder-only Transformer using the configuration in Table~\ref{tab:qwen_scales}.
As shown in Figure~\ref{fig:token_lr} (left), \sys consistently achieves lower validation loss than Muon, suggesting that dynamic spectral shaping transfers across decoder-only architectures.

\noindent\textbf{Training Token Scale.}
To evaluate robustness across training budgets, we vary the number of training tokens from 2.5B to 20B for the 601M model.
As shown in Figure~\ref{fig:token_lr} (middle), \sys consistently achieves lower validation loss than Muon across all tested budgets, suggesting that the benefit of dynamic spectral shaping is not tied to a particular training horizon.

\noindent\textbf{Learning Rate.}
We test learning-rate robustness on the 127M model under the 10B-token budget by sweeping Muon and \sys over learning rates from $0.003$ to $0.04$.
As shown in Figure~\ref{fig:token_lr} (right), \sys outperforms Muon across all tested learning rates and has a flatter curve near its optimum, indicating lower sensitivity to learning-rate choice.

\noindent\textbf{Training Dataset.}
We test corpus robustness by replacing FineWeb with FineWeb-Edu on the 127M model while keeping all other settings unchanged.
As shown in Figure~\ref{fig:data_pmin} (left), \sys consistently outperforms Muon on FineWeb-Edu, with the advantage becoming more pronounced in the late stage.
This suggests that the benefit of dynamic spectral shaping is robust to changes in the training corpus.

\noindent\textbf{Ablation on $p_{\min}$.}
We vary the scheduling endpoint $p_{\min}$ on the 127M model.
As shown in Figure~\ref{fig:data_pmin} (middle), mildly negative choices outperform Muon, with $p_{\min}=-0.25$ achieving the best validation loss.
More aggressive negative choices, e.g., $p_{\min}=-0.5$, perform worse, consistent with our analysis that overly negative exponents can degrade training performance.

\noindent\textbf{Comparison with Exact Spectral Operations.}
We compare \sys with an exact-SVD implementation of the same dynamic spectral schedule.
As shown in Figure~\ref{fig:data_pmin} (right), \sys closely matches exact SVD in validation loss, and both outperform Muon.
Since exact SVD is roughly $3\times$ slower, \sys captures the benefit of dynamic spectral shaping at much lower cost.

\noindent\textbf{Ablation on Spectral Scheduling, Logistic Parameters $(\tau,w)$.}
We ablate both the scheduling strategy and the logistic schedule parameters $(\tau,w)$.
As shown in Figure~\ref{fig:ablation} (left), our default logistic schedule outperforms standard Muon and two ablations: an abrupt switch from $p=1$ to $p=-0.25$ at step 500, and a fixed negative schedule with $p=-0.25$ throughout training.
The abrupt schedule underperforms the logistic schedule, suggesting that a smooth transition between spectral shaping is more effective than a sharp switch.
The fixed negative schedule performs substantially worse, showing that negative shaping throughout training is insufficient and supporting our stage-dependent design.
Figures~\ref{fig:ablation} (middle, right) further show that \sys is reasonably robust to the transition point $\tau$ and transition width $w$, with the best performance observed around $\tau=0.04$ and $w=0.04$. 

\noindent\textbf{Additional Results}
Additional results in~\Cref{app:results} show that \sys consistently outperforms Muon across random seeds with low variance, and also improves over NorMuon.

\noindent\textbf{Robustness Across Loss Objectives.}
In~\Cref{app:discuss}, we evaluate on a larger family of the CE--Brier loss interpolations~\citep{glenn1950verification}, to test whether mild late-stage negative shaping remains beneficial beyond standard Cross-Entropy (CE).
Interestingly, we find that across these objectives, mildly negative exponents also consistently improve over the corresponding $p=0$ baseline, while overly negative exponents degrade performance, suggesting that this benefit extends across probability-space losses.

\section{Conclusions and Future Directions}\label{sec:conclusion}
We studied a broader power-law spectral shaping family, $U\Sigma^pV^\top$. 
Our noise-aware local model reveals a stage-dependent signal-noise trade-off: early training benefits from positive $p$ that emphasizes high-curvature directions, while late training benefits from mildly negative $p$ that reallocates update strength towards flat directions. 
We proposed \sys, which efficiently schedules $p$ from positive to mildly negative values during training. 
Experiments across model scales, architectures, and training settings show that \sys consistently achieves lower validation loss than Muon.
Our experiments focus on validating the hypothesis that practical improvements are obtainable by varying $p$ over training. 
We show that this schedule outperforms fixed $p$ choices and the standard $p=0$ Muon update, but it is not the final word on the matter. 
Indeed, the best choice of $p$ may depend on training-time dynamics that cannot be fully anticipated in advance.
This suggests interesting directions for future work: selecting $p$ \emph{online} based on observed optimization statistics.
Such methods could use more expensive signals, provided they are applied only occasionally rather than at every training step.
Since the useful range of $p$ appears relatively small, we conjecture that a future near-optimal adaptive Muon-style scheduler may only need to re-estimate $p$ occasionally during training.

\newpage

\bibliographystyle{plain}  
\bibliography{src/references} 

\newpage


\appendix

\begin{table}[t]
\centering
\small
\begin{tabular}{lcccc}
\toprule
\textbf{Model} & \textbf{$d_{\mathrm{model}}$ / Layers / Heads} & \textbf{Tokens/Step} & \textbf{Total Steps} & \textbf{Total Tokens} \\
\midrule
127M  & 512 / 24 / 8    & 0.524M & 20K & 10B \\
601M  & 1280 / 24 / 20  & 0.524M & 20K & 10B \\
1.11B & 1792 / 24 / 28  & 0.524M & 20K & 10B \\
\bottomrule
\end{tabular}
\caption{Model scales for the GPT-style architecture used in our main experiments.}
\label{tab:gpt_scales}
\end{table}

\begin{table}[t]
\centering
\small
\setlength{\tabcolsep}{4pt}
\begin{tabular}{lccccc}
\toprule
\textbf{Model} & \textbf{$d_{\mathrm{model}}$ / Layers / Heads / KV Heads} & \textbf{MLP Dim.} & \textbf{Tokens/Step} & \textbf{Steps} & \textbf{Total Tokens} \\
\midrule
171M & 512 / 24 / 8 / 2 & 2816 & 0.524M & 20K & 10B \\
\bottomrule
\end{tabular}
\caption{Model scale for the Qwen-style architecture used in our experiments.}
\label{tab:qwen_scales}
\end{table}

\begin{figure*}[t]
    \centering
    \includegraphics[width=0.5\linewidth]{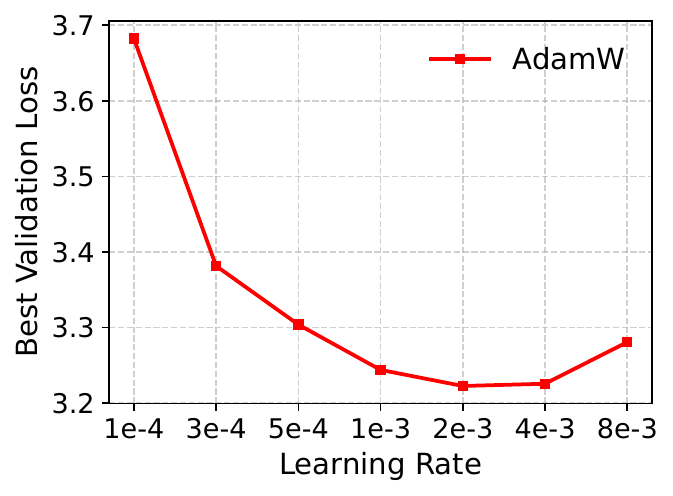}
    \caption{
    AdamW learning-rate sweep on the 127M GPT-style model, with the best validation loss achieved at $2\times 10^{-3}$.
    }
    \label{fig:adamw_lr}
\end{figure*}

\begin{figure*}[t]
    \centering
    \includegraphics[width=\linewidth]{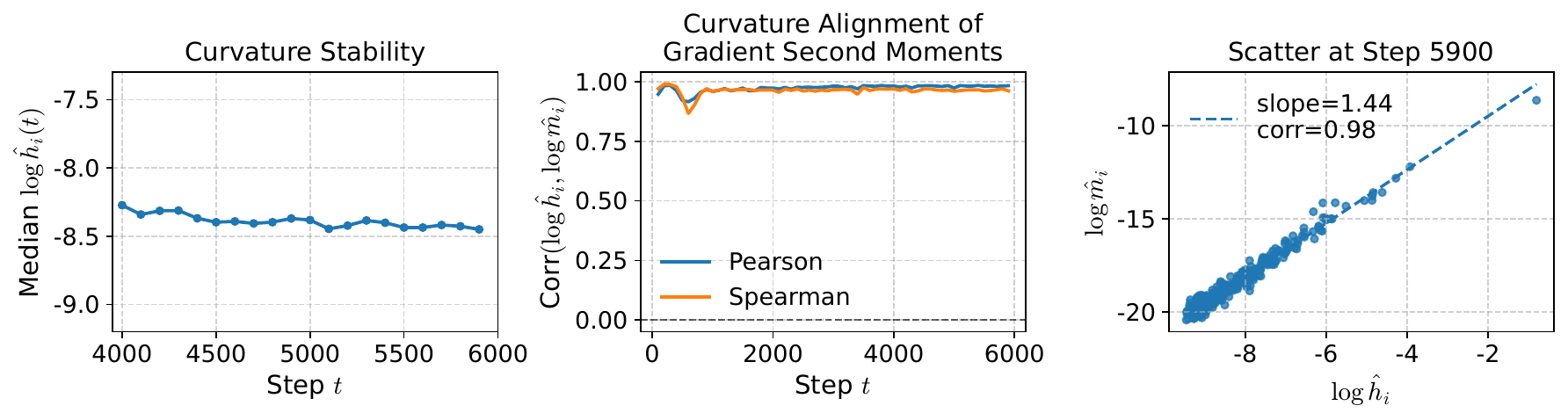}
    \caption{
        Empirical support for curvature stability and gradient-curvature alignment.
        \textbf{Left:} The median of $\log \hat{h}_{i,t}$ over the retained empirical modes changes only mildly within the 4k--6k step window, supporting our approximation that the effective local curvature can be treated as approximately fixed over short windows.
        \textbf{Middle:} Pearson ($\uparrow$) and Spearman ($\uparrow$) correlations across retained empirical modes between the local curvature proxy $\log \hat{h}_{i,t}$ and the gradient second-moment proxy $\log \hat{m}_{i,t}$, where $\hat{m}_{i,t}:=\hat{c}_{i,t}+(g^{\mathrm{probe}}_{i,t})^2$, remain strongly positive throughout training.
        \textbf{Right:} At a representative step ($t=5900$), the scatter plot of $\log \hat{m}_{i,t}$ versus $\log \hat{h}_{i,t}$ shows a clear positive relationship.
        The left panel supports the local-stability assumption for the effective curvature. 
        The middle and right panels support the gradient-curvature alignment used in the curvature-based approximation to the spectral preconditioner in~\Cref{eq:matrix_error_dynamics_model}.
    }
    \label{fig:grad2-curv-alignment}
\end{figure*}

\begin{figure*}[t]
    \centering
    \includegraphics[width=0.95\linewidth]{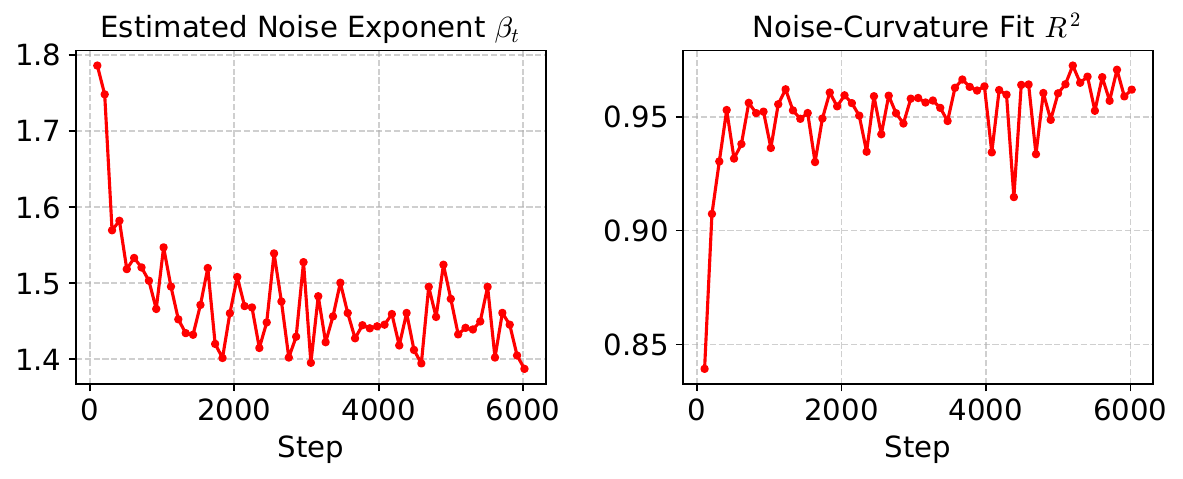}
    \caption{
    Trends in the estimated noise exponent $\beta_t$ and the noise-curvature fit $R^2$ during training.
    The power-law relationship between noise and curvature remains stable and pronounced throughout training.
    }
    \label{fig:beta}
\end{figure*}

\begin{figure*}[t]
    \centering
    \includegraphics[width=\linewidth]{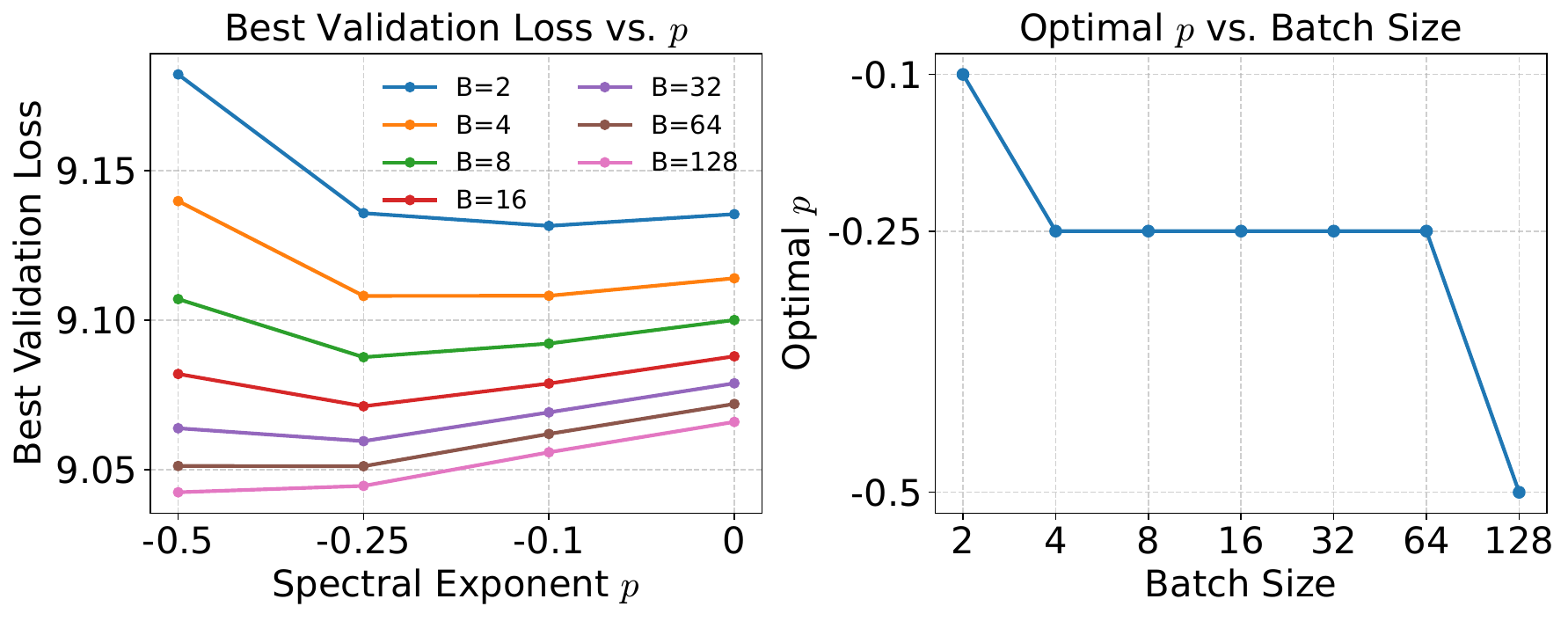}
    \caption{
    Impact of batch size on the preferred spectral exponent $p$.
    Left: best validation loss as a function of $p$ under different batch sizes.
    Right: the optimal $p$ selected by the best validation loss for each batch size.
    Smaller batch sizes induce higher gradient noise and favor mildly negative values of $p$ closer to $0$, consistent with our analysis that noise amplification limits the benefit of overly negative spectral exponents.
    }
    \label{fig:noise_impact}
\end{figure*}

\begin{figure*}[t]
    \centering
    \includegraphics[width=\linewidth]{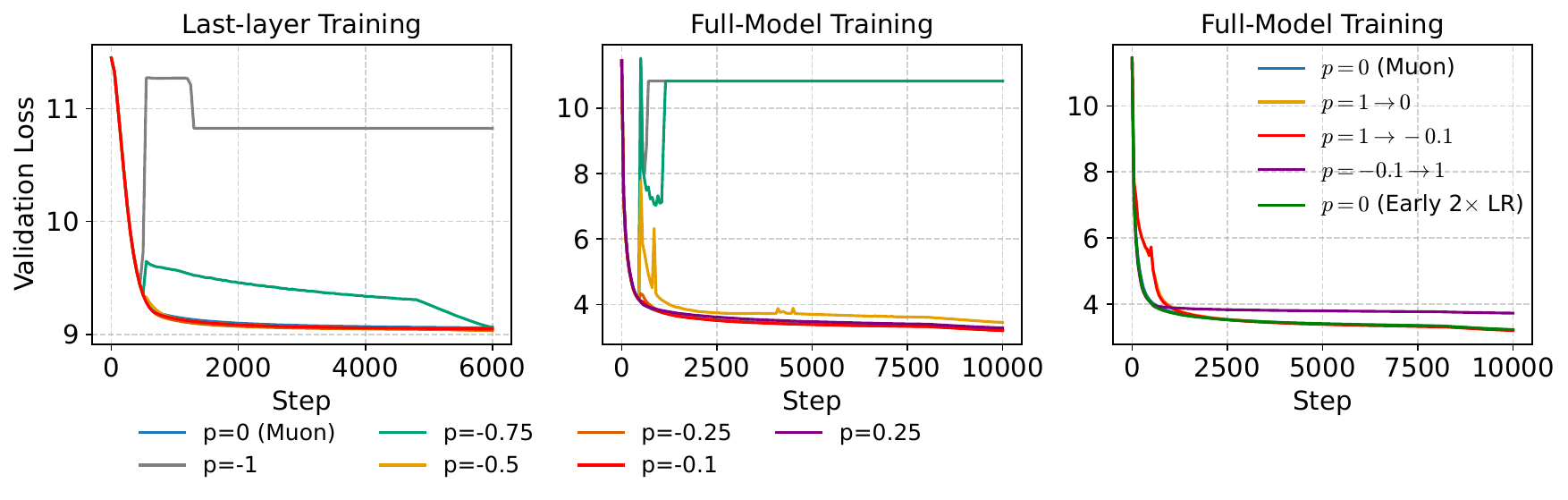}
    \caption{
        Training performance of stage-dependent spectral shaping.
        \textbf{Left/Middle:} Mildly negative exponents improve late-stage validation loss in both last-layer and full-model training, while overly negative and late-positive exponents degrade performance.
        \textbf{Right:} Early positive $p$ improves full-model training, with the positive-to-negative schedule achieving the lowest validation loss. 
        In contrast, simply doubling the early learning rate or reversing the schedule from negative to positive performs worse than Muon.
    }
    \label{fig:train_obs_full}
\end{figure*}

\begin{figure*}[t]
\centering
\begin{minipage}[t]{0.49\textwidth}
\begin{algorithm}[H]
\caption{
\tsf{\sys}
}
\label{alg:dyn-full}
\begin{algorithmic}[1]
\STATE \textbf{Input:} Update matrix $M$, step $t$, total steps $T$, schedule parameters $(p_{\max}, p_{\min}, \tau, w)$

\STATE {\ttfamily \textcolor{red}{\(\triangleright\) /* Logistic Scheduling */}}
\STATE $u \gets \left({t}/{T}-\tau\right)/w $
\STATE $ a \gets 1/(1+\exp(u))$
\STATE $p_t \gets p_{\min}+a(p_{\max}-p_{\min})$

\STATE {\ttfamily \textcolor{red}{\(\triangleright\) /* Positive Anchoring */}}
\IF{$p_t \geq 1/4$}
    \RETURN $M$
\ELSIF{$p_t \geq 0$}
    \RETURN \texttt{Newton--Schulz}$(M)$
\ELSE \RETURN \tsf{Fast--Spectral}$(M, p_t)$
\ENDIF
\end{algorithmic}
\end{algorithm}
\end{minipage}
\hfill
\begin{minipage}[t]{0.49\textwidth}
\begin{algorithm}[H]
\caption{
\tsf{Fast--Spectral}
}
\label{alg:efficient}
\begin{algorithmic}[1]
\STATE \textbf{Input:} Target matrix $X$, spectral exponent $p$
\STATE $X_n \gets X/ \|X\|_F$,  
\STATE {\ttfamily \textcolor{red}{\(\triangleright\) /* Compute Muon Update */}}
\STATE $Y_{\mu} \gets \tsf{Newton--Schulz}(X_n)$
\STATE {\ttfamily \textcolor{red}{\(\triangleright\) /* Low-Order Correction */}}
\STATE $A \gets X_n X_n^\top $
\STATE $E \gets A-I$
\STATE $\delta \gets p/2$
\STATE $C \gets I + \delta E + \frac{1}{2}\delta(\delta-1)E^2$
\STATE {\ttfamily \textcolor{red}{\(\triangleright\) /* Rescaling */}}
\STATE $\widetilde{X} \gets \|X\|_F^{p} C Y_{\mu}$
\RETURN $\widetilde{X}$
\vspace{0.75\baselineskip}
\end{algorithmic}
\end{algorithm}
\end{minipage}
\end{figure*}

\begin{figure*}[t]
    \centering
    \includegraphics[width=0.5\linewidth]{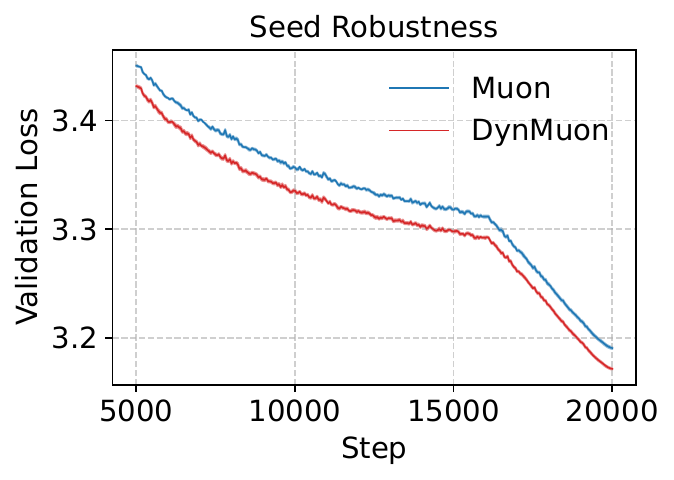}
    \caption{
    Mean validation loss across three random seeds.
    Shaded regions indicate one standard deviation, showing that \sys consistently outperforms Muon with very low seed variability.
    }
    \label{fig:std}
\end{figure*}

\begin{figure*}[t]
    \centering
    \includegraphics[width=0.5\linewidth]{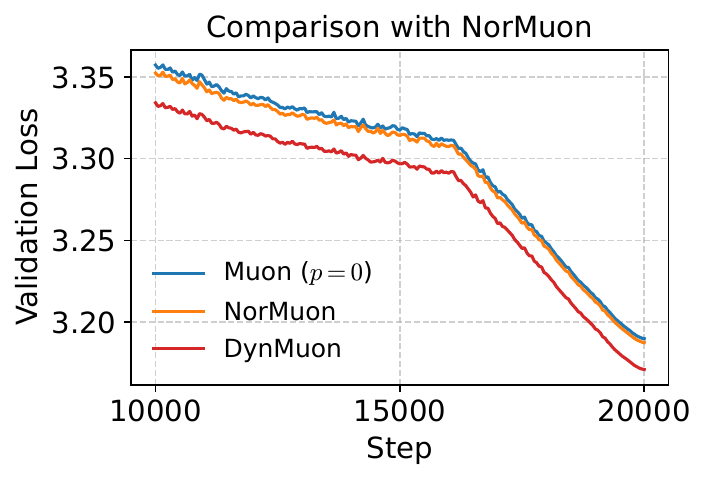}
    \caption{
        Comparison with NorMuon on the 127M model.
        \sys outperforms both Muon and NorMuon in validation loss.
    }
    \label{fig:normuon}
\end{figure*}

\begin{figure*}[t]
    \centering
    \includegraphics[width=0.5\linewidth]{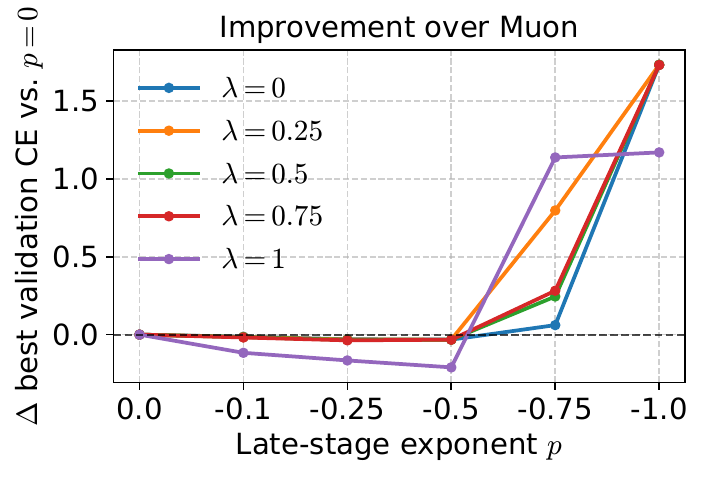}
    \caption{
        Robustness of mild negative spectral shaping across loss objectives.
        We plot the best validation CE relative to the corresponding $p=0$ baseline for each $\lambda$.
        Negative values indicate improvement over Muon.
        Mildly negative exponents remain beneficial across the CE--Brier interpolation, whereas overly negative exponents degrade performance.
    }
    \label{fig:discuss}
\end{figure*}

\section{Validation Details in~\Cref{sec:obs}}\label{app:obs_exp}
\subsection{Validation Setup}
To validate the predictions of our model, we run experiments on a GPT-style model with hidden dimension 768, 12 layers, and 6 attention heads~\citep{ahn2025diondistributedorthonormalizedupdates}, trained on FineWeb~\citep{NEURIPS2024_370df50c}.
We consider two settings: last-layer-only training and full-model training.
For last-layer-only training, we freeze all model parameters and optimize only one selected matrix-valued parameter in the final Transformer block.
We train for 6000 steps in the last-layer-only setting and 10000 steps in the full-model setting, using up to 3B training tokens.
For optimization, we use a learning rate of $0.01$ for matrix-valued parameter updates under both Muon and stage-wise spectral shaping.
For non-matrix parameter groups, we use AdamW with a learning rate of $0.001$.
We apply a weight decay of $0.01$ to the main matrix parameter groups.
The learning rate uses a linear warmup for the first $1\%$ of training steps, followed by cosine decay with a final warmdown ratio of $0.2$.
For these validation experiments, we use exact SVD to compute spectral shaping for different exponents.
For the experiments in~\Cref{sec:negative}, we switch from $p=0$ to a negative exponent at step 500 and compare $p \in \{-0.1,-0.25,-0.5,-0.75,-1\}$ in both last-layer-only and full-model training.
For the validations in~\Cref{sec:positive}, we use a simple two-stage schedule that sets $p=1$ for the first 500 steps and then switches to either $p=0$ or $p=-0.1$.

\subsection{Empirical Modes and Proxies}
In our analysis, modes are curvature directions, which are expensive to track directly in practice.
We therefore use the singular directions of the Muon update as empirical modes, since spectral shaping reweights singular values along these directions.
For efficiency, at each step $t$, we retain the top-$k$ directions with $k=256$ and represent each as
$B_{i,t}:=u_{i,t}v_{i,t}^\top$.

For each $B_{i,t}$, we estimate the empirical local curvature using a Hessian-vector product: 
\begin{equation*}
    \hat{h}_{i,t} = \langle \nabla^2 L(W_t)  B_{i,t}, B_{i,t}\rangle,
\end{equation*}
This gives the curvature scale along the empirical mode $B_{i,t}$.

We estimate the mode-wise noise level by measuring how much the gradient projection varies across independent mini-batches.
Specifically, for $n_b=32$ mini-batches, we compute
\begin{equation*}
g_{i,t}^{(b)} := \langle G_t^{(b)}, B_{i,t}\rangle,
\end{equation*}
and define the noise level proxy as the sample variance
\begin{equation*}
\hat{c}_{i,t} := \mathrm{Var}_{b\in[n_b]} [g_{i,t}^{(b)}].
\end{equation*}

To estimate residual signal energy, we use a separate fixed probe set of 8 mini-batches.
Let $g^{\mathrm{probe}}_{i,t}$ denote the gradient projection on this fixed probe set.
Since the probe gradient is averaged over a fixed batch set, we use it as a low-variance proxy for the population-gradient projection along $B_{i,t}$.
Using the local relation $g_{i,t} \approx \kappa_t h_i\delta_{i,t}$, we estimate 
\begin{equation*}
    \hat{\delta}_{i,t}^{2} := (g^{\mathrm{probe}}_{i,t})^{2} / \hat{h}_{i,t}^{2},
\end{equation*}
Here, $\hat h_{i,t}$ is the HVP-based empirical curvature and already includes the local curvature scale.

\section{Additional Analysis in~\Cref{sec:obs}}\label{app:noise}
\noindent\textbf{Stability of Effective Directional Curvatures.}
To examine the local-stability assumption in~\Cref{eq:local_approx}, we track
$\operatorname{median}_{i\in[k]}\log \hat{h}_{i,t}$ over the 4k--6k step window,
where $\hat{h}_{i,t}$ denotes the empirical local curvature estimated along the corresponding direction $B_{i,t}$, as defined in~\Cref{sec:validation}.
As shown in Figure~\ref{fig:grad2-curv-alignment} (left), this median changes only mildly over this window, supporting our use of an approximately fixed effective local curvature over short training windows.

\noindent\textbf{Gradient Second Moments Align with Curvature.}
To support the curvature-alignment approximation in~\Cref{sec:modeling}, we compare $\hat{m}_{i,t}$ with $\hat{h}_{i,t}$ over the retained empirical modes.
For each retained empirical mode $i$, define
\begin{equation*}
    \hat{m}_{i,t}:=\hat{c}_{i,t}+(g^{\mathrm{probe}}_{i,t})^2,
\end{equation*}
where $\hat{c}_{i,t}$ estimates the mode-wise noise level and $g^{\mathrm{probe}}_{i,t}$ estimates the population-gradient projection.
Thus, $\hat{m}_{i,t}$ estimates the gradient second moment along that empirical mode.
At each saved step, we compute Pearson and Spearman correlations over the retained empirical modes between $\log \hat{m}_{i,t}$ and $\log \hat{h}_{i,t}$.
As shown in Figure~\ref{fig:grad2-curv-alignment} (middle), these correlations remain strongly positive throughout training, providing empirical support for the approximation that gradient second moments are aligned with the local effective curvature.
Figure~\ref{fig:grad2-curv-alignment} (right) further illustrates this alignment at a representative step ($t=5900$), showing a strong positive relationship between $\log \hat{m}_{i,t}$ and $\log \hat{h}_{i,t}$ over the retained empirical modes.

\noindent\textbf{Noise-Curvature Scaling.}
Motivated by prior work on the geometry of gradient noise, we examine how noise level varies with curvature across modes by fitting a power-law relation $\hat{c}_{i,t} \asymp N_t \hat{h}_{i,t}^{\beta_t}$.
The exponent $\beta_t$ describes the curvature dependence of noise, where a positive $\beta_t$ means that the raw noise level is larger on high-curvature modes and smaller on flat modes.
Figure~\ref{fig:beta} (left) shows that $\beta_t$ remains stable around $1.4$, while Figure~\ref{fig:beta} (right) shows consistently high $R^2$ values, indicating that this power-law relation provides a reliable description of the mode-wise noise structure.
Thus, although decreasing $p$ amplifies noise more strongly on flat modes, their raw noise level remains comparatively smaller, leaving room for slightly negative spectral shaping to exploit the flat-mode residual signal.

\noindent\textbf{Impact of Gradient-Noise Amplification on the Preferred Spectral Exponent $p$.}
To investigate how gradient-noise amplification affects the preferred spectral exponent $p$ in the late stage, we vary the batch size from $2^1$ to $2^7$ to induce different noise levels.
We compare negative spectral exponents $p\in\{-0.1,-0.25,-0.5\}$ with Muon ($p=0$) in last-layer training.
As shown in Figure~\ref{fig:noise_impact}, smaller batch sizes, which induce higher gradient noise, favor mildly negative exponents closer to $0$: the best exponent is $p=-0.1$ when the batch size is $2$, and shifts to $p=-0.25$ when the batch size is $16$.
As the batch size further increases to $128$, the preferred exponent becomes more negative, with $p=-0.5$ achieving the best validation loss.
This trend is consistent with our analysis: negative spectral shaping can improve late-stage optimization by emphasizing flat modes, but overly negative exponents also amplify noise and can degrade performance, especially when the gradient-noise level is high.
These results also suggest that reducing gradient noise, for example, through larger batch sizes, may make more negative spectral exponents beneficial.

\section{\sys Algorithm Details}\label{app:algo}
We provide the full and efficient implementations of \sys in~\Cref{alg:dyn-full,alg:efficient}.
To facilitate reproducibility, we release a lightweight implementation of these core algorithmic components, including the spectral-exponent schedule and the fast spectral-shaping approximation.
The implementation is modular and plug-and-play: it only requires a matrix-valued update as input, applies the scheduled spectral-shaping transform, and returns the shaped update for use in existing training code.

\section{Experimental Details}\label{app:exp_set}
\noindent\textbf{Models.}
We consider two decoder-only Transformer families: a GPT-style architecture following the modded-nanoGPT setup~\cite{ahn2025diondistributedorthonormalizedupdates}, and a Qwen-style architecture.
The GPT-style models use a GPT-2 tokenizer with a vocabulary size of $50304$, rotary position embeddings~\cite{su2024roformer}, non-parametric RMSNorm, bias-free linear layers, and squared ReLU activations in the MLP blocks~\cite{NEURIPS2021_2f3c6a4c}.
The Qwen-style models use pre-normalized residual Transformer blocks with grouped-query attention, rotary position embeddings, and a gated MLP with SiLU activation, followed by a final RMSNorm layer and a bias-free output head.
As shown in~\Cref{tab:gpt_scales,tab:qwen_scales}, we evaluate three GPT-style model scales and one Qwen-style model scale.
The GPT-style models use $d_{\mathrm{model}}=512, 1280,$ and $1792$, while the Qwen-style model uses $d_{\mathrm{model}}=512$.
Across all models, we use the same training setup, with a global batch size of $512$, per-device batch size $64$, and sequence length $1024$.

\noindent\textbf{Baselines.}
We compare \sys against two baselines: Muon~\cite{jordan2024muon} and AdamW~\cite{loshchilov2018decoupled}.
For Muon and \sys, we use a default learning rate of $0.01$, and for AdamW, we use a default learning rate of $0.002$.
We use a weight decay of $0.01$ for the main matrix parameter groups.
Since Muon and \sys are designed for matrix-valued parameter updates, we use AdamW as the default scalar optimizer with a learning rate of $0.001$ for parameters outside the main matrix groups, including the embedding table and output head.
We tune the learning rate over $\{0.003, 0.005, 0.01, 0.02, 0.04\}$ for Muon and \sys, and over $\{10^{-4}, 3\times 10^{-4}, 5\times 10^{-4}, 10^{-3}, 2\times 10^{-3}, 4\times10^{-3}, 8\times10^{-3}\}$ for AdamW.
As shown in Figure~\ref{fig:adamw_lr}, AdamW achieves its best validation loss at learning rate $2\times10^{-3}$.
We use this tuned AdamW baseline when comparing against Muon and \sys in the main experiments.
For \sys, we use $w=0.01$, $\tau=0.02$, $p_{\max}=1$, and $p_{\min}=-0.25$ by default.
For all methods, we use a linear warmup for the first $0.01$ of training steps, followed by a cosine decay over the remaining steps, with a final warmdown ratio of $0.2$.

\noindent\textbf{Dataset.}
We train our models on pretraining datasets FineWeb and FineWeb-Edu~\cite{NEURIPS2024_370df50c}.
Unless otherwise specified, our main setting uses a training budget of 10B tokens of the FineWeb dataset.
To study the effect of data scale, we vary the training budget from 2.5B to 20B tokens, corresponding to 5K, 10K, 20K, and 38K training steps, respectively.

\noindent\textbf{Devices.}
We use NVIDIA H200 GPUs for all experiments.

\section{Additional Experiments}\label{app:results}

\noindent\textbf{Seed Robustness.}
To assess the robustness of \sys to training randomness, we run Muon and \sys with three different seeds $\{0,1,42\}$.
Figure~\ref{fig:std} reports the mean validation loss with one-standard-deviation bands across seeds.
\sys consistently outperforms Muon, and the very small across-seed variance suggests that the improvement is robust to training randomness rather than seed-specific effects.

\noindent\textbf{Comparison with NorMuon.}
We further compare \sys with NorMuon~\cite{li2025normuonmakingmuonefficient}, a recent Muon variant that augments Muon orthogonalization with neuron-wise normalization based on second-moment statistics.
We implement NorMuon following its original algorithmic design in our controlled setting.
As shown in Figure~\ref{fig:normuon}, \sys consistently achieves lower validation loss than both Muon and NorMuon.

\section{Discussion on Robustness Across Loss Objectives}\label{app:discuss}
The preferred spectral exponent may depend not only on the optimizer but also on the loss objective, since different losses induce different gradient and noise structures.
Cross-Entropy (CE) emphasizes poorly predicted target tokens through the $-\log \hat y_c$ term, whereas squared-error-like probability losses impose smoother penalties on probability errors.
These differences can alter the distribution of the residual signal and stochastic noise across spectral directions.
Since $p$ controls the relative update strength across these directions, we investigate whether the mild negative regime identified above is specific to CE or remains beneficial under other probability-space objectives.

To investigate this question, we use a simple parameterized family of probability-space losses that interpolates between CE and a squared-error-like objective:
\begin{equation}
L_\lambda(y,\hat y) = (1-\lambda)\mathrm{CE}(y,\hat y) + \lambda \mathrm{Brier}(y,\hat y),
\end{equation}
where $\hat y$ denotes the predicted probability distribution and $\mathrm{Brier}(y,\hat y)=\sum_i (y_i-\hat y_i)^2$ is the squared error on predicted probabilities~\citep{glenn1950verification}.
The endpoint $\lambda=0$ recovers the standard CE objective, while $\lambda=1$ gives the Brier score, a probability-space analogue of MSE.
By varying $\lambda$, we systematically change how the loss weights prediction errors and test whether the preferred mild-negative spectral regime remains robust.

We use the same last-block target-matrix setting as in~\Cref{sec:negative}, but train for 3000 steps.
For each loss interpolation parameter $\lambda\in\{0,0.25,0.5,0.75,1.0\}$, we sweep late-stage exponents $p\in\{0,-0.1,-0.25,-0.5,-0.75,-1.0\}$.
As before, we use $p=0$ before the switch step and apply the target exponent afterward.
Because the training objective changes with $\lambda$, objective values are not directly comparable across different loss choices.
We therefore evaluate all runs using validation CE as a common metric, since it is the standard language-modeling measure of next-token predictive quality and is directly tied to perplexity~\citep{kaplan2020scalinglawsneurallanguage}.

For each configuration $(\lambda,p)$, we report the difference between its best validation CE and that of the corresponding $p=0$ baseline under the same $\lambda$.
Results in~\Cref{fig:discuss} show that mildly negative exponents consistently improve upon the Muon baseline across the CE--Brier interpolation, while more aggressive negative exponents such as $p=-0.75$ and $p=-1$ degrade performance.
This suggests that the late-stage benefit of mild negative spectral shaping is not specific to standard cross-entropy, but persists across this family of probability-space loss objectives.

\noindent\textbf{Parallel Training.}
\sys preserves the matrix-wise update structure of Muon and only changes the local spectral shaping rule through the scheduled exponent $p_t$. 
It therefore introduces no additional cross-layer or cross-device coupling beyond Muon, and should be compatible with existing Muon-style parallel training implementations.

\section{Extended Related Work}\label{app:related}
\noindent\textbf{Orthonormalization-Based Optimizers for LLM Training.}
Recent work has shown that explicitly orthonormalizing matrix-shaped momentum can substantially improve optimization in large-scale neural network training~\citep{jordan2024muon,liu2025muonscalablellmtraining,wen2026fantastic,semenov2025benchmarkingoptimizerslargelanguage}. 
In particular, Muon applies Newton--Schulz-style iterations to transform matrix-valued momentum into an orthonormalized update direction, and was introduced as a structure-aware optimizer for hidden-layer matrices with strong empirical performance in neural network training~\citep{jordan2024muon}. 
Subsequent large-scale studies further demonstrate that Muon remains effective in language model pretraining when combined with appropriate weight decay and update scaling, supporting orthonormalization-based optimization as a practical paradigm beyond small-scale settings~\citep{liu2025muonscalablellmtraining}. 
Beyond empirical scaling, recent theory has also begun to explain why orthonormalization-based updates can be effective, interpreting Muon through non-Euclidean trust-region optimization, norm-constrained optimization, and spectral preconditioning viewpoints~\citep{kovalev2025understandinggradientorthogonalizationdeep,pethick2025training,chen2026muon,ma2026preconditioningbenefitsspectralorthogonalization}. 
Taken together, these works establish orthonormalization as a promising matrix-aware optimization principle and motivate studying the spectral design of matrix-valued updates more directly.

\noindent\textbf{Spectral Shaping Beyond Fixed Muon Updates.}
Recent work has begun to view Muon through a broader spectral lens, rather than treating it solely as a fixed orthonormalization rule. 
For instance, \cite{qi2026delvingmuonbeyonddeep} places Muon within a family of spectral operators of the form $U\Sigma^pV^\top$ and studies how different fixed choices of positive $p$ connect Muon-style updates to momentum and Adam-like normalization. 
Other recent extensions further explore richer but still largely fixed or task-specific forms of spectral shaping. 
For example, Spectra~\citep{huang2026spectrarethinkingoptimizersllms} argues that LLM training exhibits persistent spectral anisotropy with a dominant spike subspace and a long informative tail, and proposes spike-aware shaping that suppresses dominant directions without amplifying the noise-sensitive tail. 
PRISM~\citep{yang2026prismstructuredoptimizationanisotropic} augments first-order spectral descent with low-rank quasi-second-order information to perform anisotropic spectral shaping, while SpecMuon~\citep{lu2026muonspectralguidanceefficient} introduces mode-wise spectral guidance for scientific machine learning settings with stiff multi-scale dynamics. 
More broadly, beyond Muon-specific extensions, matrix-aware optimizers such as Shampoo, SOAP, and PolarGrad also exploit non-diagonal geometry and spectral structure to reshape updates beyond simple coordinate-wise scaling~\citep{pmlr-v80-gupta18a,shi2023distributeddataparallelpytorchimplementation,vyas2025soap,lau2026polargradclassmatrixgradientoptimizers}. 
These works establish spectral structure as an important design axis for matrix-valued optimization. 
However, they mainly study fixed or task-specific spectral transformations. 
In contrast, \sys asks whether the preferred spectral shaping should evolve across different training stages, and studies dynamic spectral shaping rather than a single fixed spectral rule.
We primarily compare against Muon because \sys directly generalizes the Muon update: Muon corresponds to the fixed exponent $p=0$, whereas \sys dynamically varies $p$ within the same spectral-shaping family. 
This provides a controlled comparison that isolates the effect of the spectral schedule while keeping the rest of the optimizer design fixed. 
Muon is also a strong and widely used baseline for LLM training, with recent work showing substantial efficiency gains. 
Matrix-preconditioned optimizers such as Shampoo, SOAP, and PolarGrad are better viewed as complementary to our study rather than as controlled baselines for isolating the effect of dynamic spectral shaping, since they also change the underlying preconditioner and update rule.

\noindent\textbf{Dynamic Scheduling in Optimization.}
A long line of work improves training by dynamically scheduling optimizer hyperparameters across different phases of optimization, reflecting the broader principle that different stages of training often benefit from different update behaviors. 
Classic examples include learning-rate annealing and restart strategies such as SGDR~\citep{loshchilov2017sgdr}, cyclical and one-cycle policies that jointly vary learning rate and momentum~\citep{smith2018disciplinedapproachneuralnetwork, smith2018superconvergencefasttrainingneural}, and warmup schedules that stabilize adaptive optimizers such as Adam in the early stage of training~\citep{Ma_Yarats_2021}. 
Such schedules remain central in modern large-batch and foundation-model training, where warmup and decay of scalar step sizes are often crucial for stable and efficient optimization~\citep{You2020Large}. 
More recent work has also begun to study iteration-dependent scaling more directly, for example, by learning online scaling matrices for gradient methods~\citep{pmlr-v291-gao25a}, while recent analyses of optimization under ill-conditioned objectives suggest that the relative importance of dominant and bulk subspaces can shift over the course of training~\citep{deng2026suspiciousalignmentsgdfinegrained}.
Unlike prior schedules that mainly modulate scalar hyperparameters, \sys dynamically schedules the spectral shape of the update itself. 
This allows the preferred spectral bias to evolve across training rather than remain fixed for the entire run.

\section{Broader Impact}\label{app:broader}
This work proposes a general-purpose optimization method for training LLMs.
Its main potential benefits are improved training efficiency and performance, which may reduce compute costs and energy usage.
As a technical contribution, the method does not pose direct societal risks.
We believe it promotes LLM training, and future work can further address potential concerns by incorporating responsible use guidelines.



\end{document}